\documentclass[letterpaper, 10 pt, journal, twoside]{IEEEtran}

\usepackage[caption=false,font=normalsize,labelfont=sf,textfont=sf]{subfig}
\usepackage{array}
\usepackage{textcomp}
\usepackage{verbatim}
\usepackage{graphicx}
\usepackage{epstopdf}
\usepackage{cite}
\usepackage{amsmath,amssymb,amsfonts, amsthm}
\usepackage{algorithmic}
\usepackage{algorithm}
\usepackage{xcolor}
\usepackage{float}
\usepackage{url}
\usepackage{multirow}
\usepackage{color}
\usepackage[hidelinks]{hyperref}
\usepackage{enumerate}
\usepackage{siunitx}
\usepackage{epstopdf}
\usepackage{pbox}
\usepackage{stfloats}
\usepackage{adjustbox}
\newtheorem{assumption}{Assumption}
\newtheorem{definition}{Definition}

\begin{document}
\title{Positive-Unlabeled Constraint Learning for Inferring Nonlinear Continuous Constraints Functions from Expert Demonstrations}

\author{Baiyu Peng$^{1}$, Aude Billard$^{1}$%
\thanks{
This work was supported by euROBIN and ERC SAHR Grant.   All codes are available at \href{https://github.com/epfl-lasa/PUCL_learning_constraint}{https://github.com/epfl-lasa/PUCL\_learning\_constraint.} } 
\thanks{$^{1}$ The authors are with the LASA, School of Engineering, EPFL (Swiss Federal Institute of Technology in Lausanne), Lausanne 1015 Vaud, Switzerland.
        ({\tt\footnotesize baiyu.peng@epfl.ch} ; {\tt\footnotesize aude.billard@epfl.ch}). Corresponding author: Baiyu Peng.}%

\thanks{This work has been published in IEEE Robotics and Automation Letters (RA-L). Please refer to the final version at \href{https://doi.org/10.1109/LRA.2024.3522756}{DOI:10.1109/LRA.2024.3522756}.}
}

\markboth{Preprint Version. December, 2024}
{Peng \MakeLowercase{\textit{et al.}}: Positive-Unlabeled Constraint Learning from Expert Demonstrations} 

\maketitle
\begin{abstract}
    Planning for diverse real-world robotic tasks necessitates to know and write all constraints. However, instances exist where these constraints are either unknown or challenging to specify accurately.
    A possible solution is to infer the unknown constraints from expert demonstration. This paper presents a novel two-step Positive-Unlabeled Constraint Learning (PUCL) algorithm to infer a continuous constraint function from demonstrations, without requiring prior knowledge of the true constraint parameterization or environmental model as existing works. We treat all data in demonstrations as positive (feasible) data, and learn a control policy to generate potentially infeasible trajectories, which serve as unlabeled data. The proposed two-step learning framework first identifies reliable infeasible data using a distance metric, and secondly learns a binary feasibility classifier (i.e., constraint function) from the feasible demonstrations and reliable infeasible data. The proposed method is flexible to learn complex-shaped constraint boundary and will not mistakenly classify demonstrations as infeasible as previous methods. The effectiveness of the proposed method is verified in four constrained environments, using a networked policy or a dynamical system policy. It successfully infers the continuous nonlinear constraints and outperforms other baseline methods in terms of constraint accuracy and policy safety. This work has been published in IEEE Robotics and Automation Letters (RA-L). Please refer to the final version at \href{https://doi.org/10.1109/LRA.2024.3522756}{DOI:10.1109/LRA.2024.3522756}.
\end{abstract}


\section{Introduction}
\IEEEPARstart{P}{lanning} for many robotics and automation tasks requires explicit knowledge of constraints, which define which states or trajectories are allowed or must be avoided \cite{garcia2015comprehensive, noothigattu2019teaching}. Sometimes these constraints are initially unknown or hard to specify mathematically, especially when they are nonlinear,  possess unknown shape of boundary or are inherent to user's preference and experience. For example, human users may determine an implicit minimum distance of the robot to obstacles based on the material of the obstacle (glass or metal), the shape and density of obstacles and their own personal preference. The robot should be able to infer such a constraint to achieve the task goal and meet human's requirements. This means an explicit constraint should be inferred somehow, e.g., from existing human demonstration sets.


\begin{figure}[!t]
\centering
\subfloat{\includegraphics[width=0.14\textwidth]{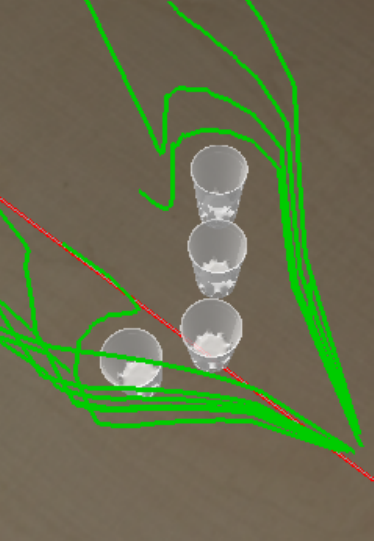}}
\hspace{\fill}
\subfloat{\includegraphics[width=0.1391\textwidth]{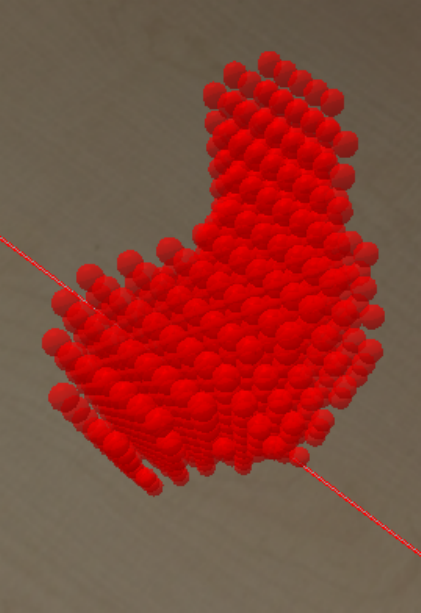}}
\hspace{\fill}
\subfloat{\includegraphics[width=0.14\textwidth]{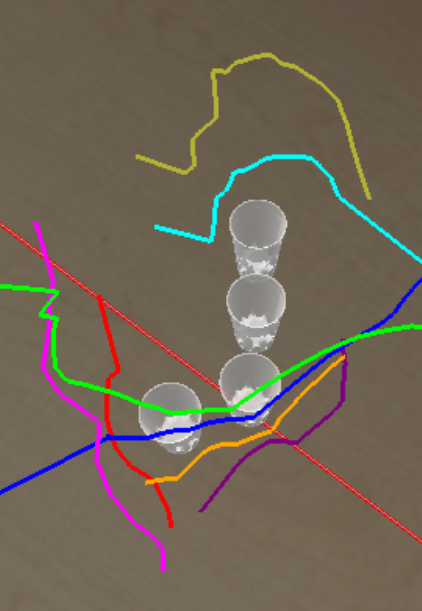}} \\
\subfloat{\includegraphics[width=0.14\textwidth]{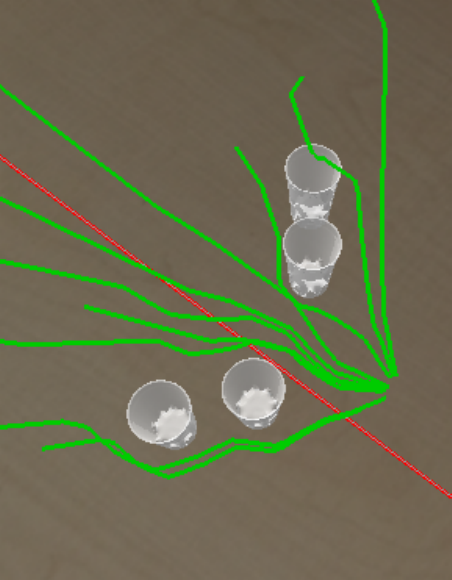}}
\hspace{\fill}
\subfloat{\includegraphics[width=0.14\textwidth]{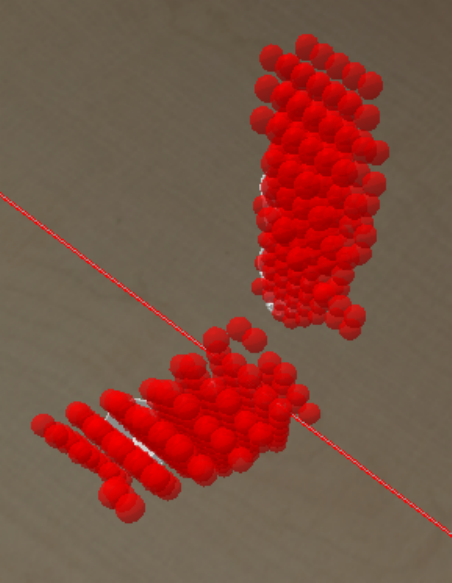}}
\hspace{\fill}
\subfloat{\includegraphics[width=0.14\textwidth]{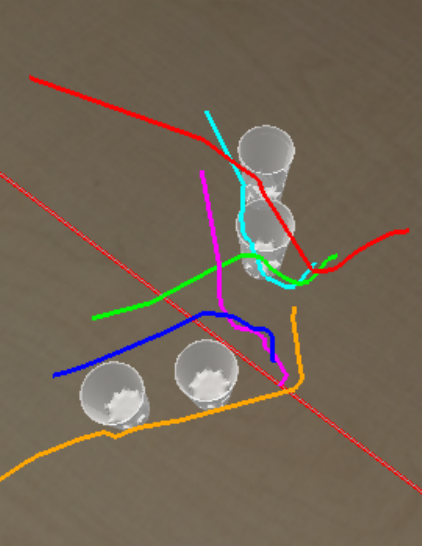}}

\caption{ 
Learning an obstacle-avoidance constraint from demonstrations with proposed method. The task requires the robot to reach a target state while avoiding going close to or over the cups. These unknown constraints can be translated into an infeasible region that must be inferred from demonstrations. The top row illustrates one obstacle configuration, while the bottom row shows another.  In each row, the left image displays the expert demonstrations (in green), the middle image shows the learned constraint region (in red), and the right image illustrates the learned policy, which is trained from the learned constraint and tested on a set of shifted goal states and starting points.  }
\label{fig:panda cups}
\end{figure}

Constraint inference from demonstrations has drawn more and more attention since 2018 \cite{chou2020learninggrid}. Various methods have been developed to learn various types of constraints. In this work, we focus on the constraints of not visiting some undesired states(-actions) throughout the trajectory (discussed in detail in section \ref{sec:preliminary}). These constraints define what the user does not want to happen and are ubiquitous in practice, e.g., not crashing into an unknown obstacle or not surpassing an unknown maximum velocity. 


Existing constraint learning methods can be roughly divided into inverse reinforcement learning (IRL)-based methods and mixed-integer-programming (MIP)-based methods.

IRL-based methods draw inspiration from the concepts and approaches of IRL and adapt them to learn the constraint function. \cite{scobeemaximum} first formulate the constraint learning problem in the maximum likelihood inference framework. They introduce a Boltzmann policy model, where the likelihood of any feasible trajectory is assumed to be proportional to the exponential return of the trajectory, while the likelihood of any infeasible trajectory is 0. Then a greedy algorithm is proposed to add the smallest number of infeasible states that maximize the likelihood of the demonstrations. This framework has the advantage of being able to work with sub-optimal demonstrations. 
Further, \cite{glazier2022learning} and \cite{mcpherson2021maximum} extend the maximum likelihood framework from a deterministic setting into a stochastic setting, with soft or probabilistic constraints. 
\cite{stocking2022maximum} proposes to learn a neural network policy via deep RL and  identify discrete infeasible states with the policy network. All aforementioned methods only extract constraint from a pre-defined finite constraint set. To learn an arbitrary continuous constraint function, \cite{anwar2020InverseCR} proposes Maximum Entropy Constraint Learning (MECL) algorithm. It not only approximates the policy with a network but also learns a constraint network, which is optimized by making a gradient ascent on the maximum likelihood objective function. More recently, \cite{liu2022benchmarking} proposes a variational approach to infer the distribution of constraint and capture the epistemic uncertainty. However, despite that \cite{anwar2020InverseCR, liu2022benchmarking} can learn a continuous constraint function, they have following shortcomings: theoretically, their loss function optimizes over all the states visited by the policy to be infeasible, and states visited by the demonstrations to be feasible. If demonstrations and policy overlap, this loss will lead to conflicts during the update step, slowing the training time and prone to classifying safe demonstrated states as infeasible. 
In addition to that, in practice  \cite{anwar2020InverseCR, liu2022benchmarking , jang2023inverse} are used to recover linear constraints, and on a unidimensional state, e.g. learn a constraint of the type $x \ge -3$, and, as we show later in the paper, perform relatively poorly when learning nonlinear constraints.  

In parallel, another line of research based on MIP has been developed. \cite{chou2020learninggrid} first studies how to infer infeasible states from a demonstration set in a grid world. They assume that all possible trajectories that could earn higher rewards than the demonstration must be constrained in some way, or the expert could have passed those trajectories. In practice, they use hit-and-run sampling to obtain such higher-reward trajectories and solve an integer program to recover the constrained states in a grid world. This MIP method is soon extended to continuous state space and parameterized continuous constraint functions, where the constraint is represented as unions and intersections of axis-aligned boxes \cite{chou2020learninglocal}. Later, the same authors propose to leverage the closed-form environmental model and the KKT optimality condition to infer the parameters of a known constraint parameterization \cite{chou2020learninglocal}, or a Gaussian Process constraint model \cite{chou2022gaussian}. However, these MIP-based methods  either require the full knowledge of a differentiable model of the environment and agent \cite{chou2020learninglocal, chou2022gaussian}, which is not always available in practice , or can only learn constrained region expressed with unions and intersections of box \cite{chou2020learningpara}. 

In summary, despite these impressive advances, it still remains to be shown how to learn continuous nonlinear constraint functions, without full knowledge of true constraint parameterization and environmental model. To overcome the aforementioned challenge, we propose a two-step Positive-Unlabeled Constraint Learning (PUCL) method,  inspired by a machine learning subarea Positive-Unlabeled (PU) learning \cite{bekker2020learning}. Within our framework, we treat all data in demonstrations as positive (feasible) dataset, and learn a policy to generate many potentially unsafe trajectories which serve as unlabeled data with unknown feasibility.  Then a two-step PU learning technique is applied, where  we first identify reliable infeasible data and secondly learn a binary feasibility classifier (i.e., constraint function) from the positive data and reliable infeasible data. The proposed method is flexible to learn complex-shaped constraint boundary without the need to know the true constraint parameterization. It also avoids the drawback of some IRL-based methods that mistakenly optimize and classify demonstrations as infeasible, thus improving the performance and accuracy. 


\section{Preliminaries and Problem Statements}
\subsection{Preliminaries and Notations}
\label{sec:preliminary}
We consider a finite-horizon deterministic Markov decision process (MDP) \cite{sutton2018reinforcement}, where the states and actions are denoted by $s \in \mathcal{S}$ and $a \in \mathcal{A}$.  $\gamma$ denotes the discount factor, and the real-valued reward is denoted by $r(s,a)$.  A trajectory $\tau = \{s_{1}, a_{1}, \ldots, s_{T}, a_{T}\}$ contains a sequence of state-action pairs in one episode. For the trajectory $\tau$, the total discounted reward (return) is defined as $r(\tau) = \sum_{t=1}^{T} \gamma^{t} r(s_{t}, a_{t})$. A policy, the mapping between states and actions, is denoted by $a = \pi(s)$. 

In this study, we focus on learning the Markovian state-action constraints, i.e., avoiding visiting some forbidden states (and/or actions) throughout the entire trajectory. Mathematically, we denote the true constraint set by $\mathcal{C^*}=\{(s,a) \in \mathcal{S}\times \mathcal{A} | (s,a) \text{ is truly infeasible} \}$. Thus, the true constraint is written as $(s,a) \notin \mathcal{C^*}$. This definition implicitly assumes that the studied constraint is  (1) time-independent; (2) deterministic: a state-action pair is either truly feasible or infeasible. In addition to the unknown constraints, some known constraints could also be present in the task, e.g., goal state constraints and system dynamic constraints. For simplicity of explanation, all demonstrations and trajectories are assumed to satisfy the known constraints (trajectories violating known constraints are discarded and not used for inferring the unknown constraint).


We will approximate the true constraint set with a constraint network $\zeta_\theta$, with $\theta$ as the parameter. To simplify the notation and allow for more flexibility, we first introduce a generalized state $\bar{s}\equiv \psi(s,a)$ as the input to the network, where $\psi(s,a)$  is a user-defined feature function that can be customized to incorporate prior knowledge about the constraint structure. For example, if the constraint is known to depend solely on the state (or actions), one could define $\psi(s, a) = s$ (or $a$). Readers should keep in mind that, in the following sections, we use the generalized state $\bar{s}$ as shorthand for $\psi(s,a)$. Following this notation, the constraint network is written as $\zeta_\theta(\bar{s}) \in [0,1]$, which induces a constraint set $\mathcal{C}_\theta=\{(s,a)| \zeta_\theta(\bar{s}) \le d \}$ with $d=0.5$ being the decision threshold.

The demonstration set $\mathcal{D}_\tau=\{\tau_i\}^{M_d}_{i=1}$ comprises trajectories demonstrated by a demonstrator. For convenience, we also introduce the set $\mathcal{D}=\{\bar{s}|(s,a)\in\tau_i\in\mathcal{D}_\tau\}$ consisting of generalized states from all the demonstrated trajectories. Similarly, we define the unlabeled data set $\mathcal{P}_\tau=\{\tau_i\}^{M_p}_{i=1}$ and $\mathcal{P}=\{\bar{s}| (s,a)\in\tau_i\in\mathcal{P}_\tau\}$.

\subsection{Problem Formulation}
As in similar works \cite{anwar2020InverseCR,chou2020learninggrid}, we require that the reward function to be known in advance. This assumption is reasonable in many robotics applications, where the reward might be shortest path, shortest time to reach the goal. We emphasize that the unknown constraints are not indicated in the known reward function, i.e., there is no negative reward for violating the unknown constraints. The real task is written as a constrained optimization problem:

\noindent\textbf{Problem 1} (Constrained optimization problem)
\begin{subequations}\label{eq:optimization_problem}
\begin{align}
  \max _{\{a_i \sim \pi\}^T_{i=0}} &\quad r(\tau) \label{eq:reward} \\
  \text{s.t.} &\quad (s_{i}, a_{i}) \notin \mathcal{C^*}, \quad \forall (s_{i}, a_{i}) \in \tau &\quad  \label{eq:unknown constraint} 
\end{align}
\end{subequations}
where $r(\tau)$ represents the reward function of trajectory $\tau$.
We make the following assumption of the demonstrations optimality and feasibility:
\begin{assumption} [Demonstrations Sub-optimality and Feasibility]
\label{assumption:demonstrations}
 Any demonstration $\tau_i \in\mathcal{D}_\tau$ is feasible and $\delta$-sub-optimal solution to the problem \ref{eq:optimization_problem}:
    \begin{enumerate}
        \item (feasible): $\tau_i$ satisfies \eqref{eq:unknown constraint} and all known constraints.
        \item ($\delta$-sub-optimal): $(1-\delta)r(\tau^*) \le r(\tau_i)$, where $\tau^*$ is the feasible optimal solution to the problem \ref{eq:optimization_problem}. $\delta\in[0, 1)$ is a coefficient of sub-optimality. 
    \end{enumerate}
\end{assumption}
In practice $\delta$ is treated as a hyper-parameter and specified according to the user's confidence of demonstration optimality. We formally define the constraint learning task as:
\begin{definition}[Constraint Learning]
    The task of  learning  constraint from demonstration is to recover the unknown true constraint set $C^*$ in Problem \ref{eq:optimization_problem} from the demonstration set $\mathcal{D}_\tau$, given the known reward function $r(\tau)$. 
\end{definition}

\section{Method}
\label{sec:method}

\subsection{Motivation: Constraint Inference as Positive-Unlabeled Learning}
\label{sec:classification}

\begin{figure}[htbp]
\centerline{\includegraphics[width=0.4\textwidth]{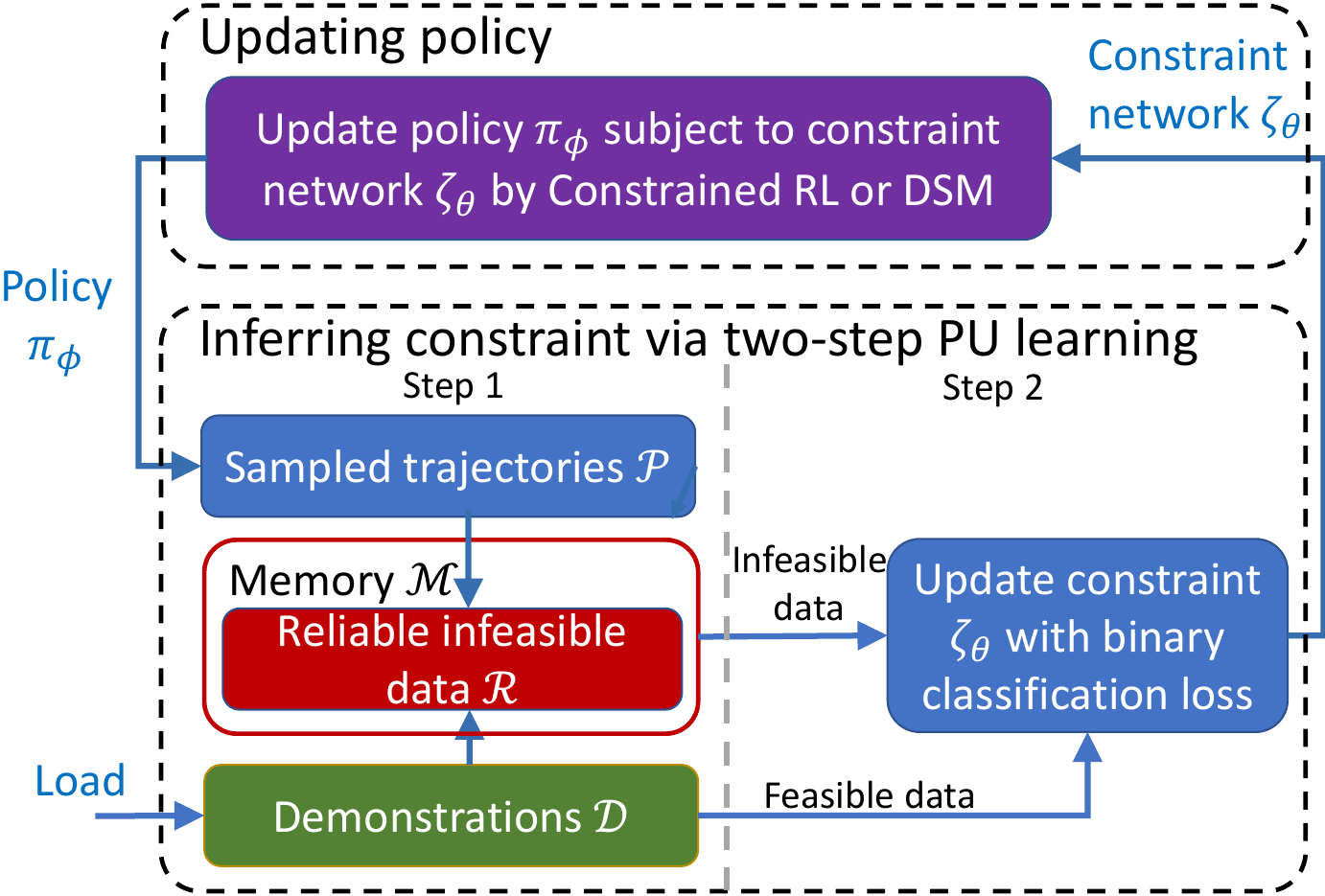}}
\caption{The framework of the proposed constraint learning algorithm PUCL. It alternates between updating policy using current constraint network, and inferring constraint from demonstrations and current policy. The constraint is inferred using a novel two-step positive-unlabeled learning technique. In the first step, the trajectories $\mathcal{P}$ sampled from current policy are viewed as unlabeled data, while the  demonstrations are viewed as labeled feasible data. From the two datasets we identify reliable infeasible data using a distance-based metric. In the second step, the reliable infeasible data from current iteration and previous iterations, as well as the feasible demonstrations, are used to train a constraint network using a standard binary classification loss. }
\label{fig:PUCL block}
\end{figure}

As in many IRL-based and MIP-based works \cite{anwar2020InverseCR, chou2020learninggrid}, we infer the underlying constraint by contrasting the (safe) demonstration with some potentially unsafe trajectories generated by a certain policy. The training follows an iterative framework to incrementally learn the constraint. In each iteration, we first sample a set of high-reward trajectories $\mathcal{P}_\tau$ by performing the current policy $\pi_\phi$ (discussed later in \ref{sec:policy learning}).
Similar to \cite{chou2020learninggrid}, we speculate that the trajectory which wins a higher reward than demonstration is potentially unsafe and cheats by visiting some forbidden states. However, it remains unclear which specific state(s) within the trajectory are unsafe. In other words, trajectory visits both truly feasible states and truly infeasible states but all states remain unlabeled. In contrast, it is certain that all states on the demonstrated trajectories are labeled feasible by Assumption 
\ref{assumption:demonstrations}.  Our goal is to classify each single state as feasible or infeasible by learning from a batch of fully labeled feasible states and another batch of unlabeled mixed states.

This insight inspires us to formulate constraint inference as a positive-unlabeled learning problem. PU learning is a machine learning technique that aims to learn a classifier from a positive dataset and an unlabeled dataset containing both positive and negative samples \cite{bekker2020learning}. Within our framework, positive samples correspond to feasible states. Thus, the data from demonstrations serve as positive data, while the policy offers unlabeled data containing some infeasible samples. We can thus utilize PU learning methods to uncover the infeasible states from the two datasets. 

PU learning methods for binary classification mainly divide into three categories: two-step methods, biased learning and weight-based methods \cite{bekker2020learning}. The latter two methods both rely on an important assumption, Sampled Completely At Random (SCAR), which requires the demonstrations and policy to have a similar distribution in the truly feasible region. This SCAR assumption could be  unrealistic and limits its application to only specific environments. In contrast, the two-step methods do not require SCAR assumption and only require the two classes can be strictly separated. Since we limit our discussion to deterministic constraint, i.e., a state is either truly feasible or infeasible, this separability assumption naturally holds. Therefore, this paper will develop a constraint learning method from a two-step framework \cite{zhang2009reliable, liu2014clustering}. 

\subsection{Method}
\label{sec:method pucl}
Our two-step PU learning method consists of the following steps: (1) identifying reliable infeasible data (denoted as $\mathcal{R}$) from unlabeled set $\mathcal{P}$, and (2) using supervised learning techniques to learn a classifier from the feasible demonstrations $\mathcal{D}$ and reliable infeasible dataset $\mathcal{R}$. 

In step 1, the identification of $\mathcal{R}$ starts from the intuition that data in the unlabeled set $\mathcal{P}$ that are very different from all data in the demonstration set $\mathcal{D}$ are highly likely to be truly infeasible. We propose a kNN-like distance-based approach, which is straightforward to comprehend and implement, to identify the reliable infeasible set $\mathcal{R}$. The unlabeled data are first ranked according to their distance to the $k$ nearest feasible data from demonstration set \footnote{Since $\mathcal{D}$ contains consecutive data points from trajectories, it is recommended to use a small $k$ around 1-3.}. Mathematically, the ranking score for each unlabeled data $\bar{s}\in\mathcal{P}$ is calculated by the average distance:
\begin{equation}
\label{score function}
    S_\mathcal{D}(\bar{s})=\frac{1}{k}\sum_{\bar{s}'\in kNN(\bar{s},\mathcal{D})} dis(\bar{s},\bar{s}') 
\end{equation}
where $kNN(\bar{s},\mathcal{D})$ indicates the $k$-nearest neighbors of $\bar{s}$ in the set $\mathcal{D}$, and $dis(\bar{s},\bar{s}')$ is a distance metric selected based on the specific task.  Then, the unlabeled data whose scores are higher than a threshold $d_r$ are selected as reliable infeasible data:
\begin{equation}
\label{RI identification}
    \mathcal{R}=\{\bar{s} \in\mathcal{P}| S_\mathcal{D}(\bar{s}) \ge d_r \}
\end{equation}
Intuitively $d_r$ indicates user's belief over the size of true infeasible region. A lower $d_r$ tends to identify a larger $\mathcal{R}$, leading to a more conservative constraint network. Conversely, a higher $d_r$ tends to find a smaller $\mathcal{R}$ and a more radical constraint function. In the extreme case of $d_r=0$, nearly any state not visited by the demonstrations is classified as infeasible. We admit selecting $d_r$ can be tricky, but such a parameter is generally inevitable since learning constraints is known to be an ill-posed problem with infinite valid solutions \cite{anwar2020InverseCR}. 

In the case where all dimensions of $\bar{s}$ are homogeneous, i.e., having similar physical meanings and units, $d_r$ will share these units and can be selected heuristically. For example, the first case in our experimental section learns an obstacle-avoidance constraint in a 3D position space and $d_r$ roughly indicates how far the demonstration is from the true obstacle. 

While in some high-dimensional case where $\bar{s}$ contains heterogeneous dimensions, identifying $\mathcal{R}$ with distance-based metric and determining the corresponding $d_r$ are more tricky. We propose to merge $\mathcal{D}$ and $\mathcal{P}$, standardize each dimension of the combined dataset to have zero mean and unit variance, and then use the standardized data to compute the scores $S_\mathcal{D}(\bar{s})$. Based on these scores, we select a percentile X and identify the top X percent of the highest-scoring data in $\mathcal{P}$ as the reliable infeasible data.

In practice, we also notice that when the demonstration closely adheres to the true constraint boundary at a short distance, kNN-like metric may fail to learn an accurate boundary, as the reliable infeasible data are all at least $d_r$ far from the demonstrations. To adapt to this case, we expand $\mathcal{R}$ by adding the state from each trajectory in $\mathcal{P}$ that is closest to the original $\mathcal{R}$, since these data are likely to lie between the current boundary and the true boundary. 
\begin{equation}
\label{RI explansion}
\mathcal{R} \leftarrow \mathcal{R} \cup ( \cup_{\tau_i \in \mathcal{P}_\tau} {\arg \min_{(s,a) \in \tau_i} S_\mathcal{R}(\bar{s})})
\end{equation}
where $S_\mathcal{R}(\bar{s})$ is the score function sharing similar definition as $S_\mathcal{D}(\bar{s})$ in \eqref{score function} but with a different set $\mathcal{R}$. Note that this expansion is  performed only once per iteration.

In step 2, we train a neural network binary classifier $\zeta_\theta(\bar{s}) $ from the reliable infeasible data $\mathcal{R}$, the memory buffer $\mathcal{M}$ (discussed later in \ref{sec:iterative framework}) and the (feasible) demonstrations data $\mathcal{D}$ with a standard cross entropy loss:
\begin{equation}
\begin{aligned}
   \mathcal{L}(\theta)=-\frac{1}{N+M}[\sum^N_{\bar{s}_i\sim \mathcal{D}} \log\zeta_\theta(\bar{s}_i)-\sum^M_{\bar{s}_j\sim \mathcal{R} \cup \mathcal{M}}\log(1-\zeta_\theta(\bar{s}_j))]
\end{aligned}
\label{BCE loss}
\end{equation}

\subsection{Iterative Learning Framework with Policy Filter and Memory Buffer}
\label{sec:iterative framework}
Last subsection discussed learning constraint from a given unlabeled set $\mathcal{P}$ and demonstrations $\mathcal{D}$. This subsection and the next  will elaborate on the generation of unlabeled set $\mathcal{P}$. Existing methods generate $\mathcal{P}$ either in an iterative manner \cite{anwar2020InverseCR, scobeemaximum, liu2022benchmarking} or in a one-batch manner \cite{chou2020learninggrid, chou2020learningpara, stocking2022maximum}. The former iteratively updates policy with current constraint function and generates new trajectories to update the constraint. Since the policy is always updated with respect to the up-to-date constraint network, the new trajectories generated in each iteration are mostly distributed around the boundary of the current constraint network, which will improve the accuracy of the constraint boundary. Therefore, this manner is efficient to learn the constraint boundary and can automatically adjust the distribution of the generated trajectories. In contrast, the one-batch manner generates a bulk of trajectories uniformly\cite{chou2020learninggrid} or greedily\cite{stocking2022maximum} at the start of training. The following training  takes place only on those trajectories, with no new data  generated. This manner is more stable but relatively inefficient and poses a higher demand on the diversity and the coverage of the generated trajectories.  

This work follows the iterative framework, and introduces a policy filter and a memory buffer.  Fig. \ref{fig:PUCL block} gives a sketch of the whole iterative structure. 

\textbf{Policy filter:} In each iteration of policy updating, the policy is updated for fixed steps with the current constraint network $\zeta_\theta(\bar{s})$. To prevent undesired updates from poor policies, 
a policy filter \eqref{filter condition} is introduced to select trajectories with a relatively high reward than the demonstration, which are believed to truly violate the unknown constraint according to Assumption \ref{assumption:demonstrations}. 
\begin{equation}
\begin{aligned}
\mathcal{P}_\tau \leftarrow \{\tau_i \in \mathcal{P}_\tau | (1-\delta)r(\tau_i) \ge r(\tau_i^\mathcal{D}) \}
\label{filter condition}
\end{aligned}
\end{equation}
where $\delta$ is a coefficient defined in Assumption \ref{assumption:demonstrations}, $\tau_i^\mathcal{D}$ is the demonstration that starts from the same state as $\tau_i$. 

\textbf{Memory buffer:}
Most iterative-framework-based papers generate new trajectories in each iteration and discard those from previous iterations \cite{anwar2020InverseCR, scobeemaximum, liu2022benchmarking}. Such a training manner tends to forget the infeasible region learned in early iterations. Thus, we introduce a memory replay buffer and record all the  reliable infeasible data $\mathcal{R}$ identified in each iteration into the memory buffer $\mathcal{M}$. In the following iterations of learning constraints, both the current $\mathcal{R}$ and memory buffer $\mathcal{M}$ serve as infeasible data for training (see \eqref{BCE loss}).

\subsection{Represent and Learn Policy via Constrained RL or Dynamical System Modulation}
\label{sec:policy learning}
In the iterative framework discussed in the last subsection, we need to maintain a policy to generate high-reward trajectories while satisfying the currently learned constraints. We consider and compare two policies: 1) constrained RL \cite{stooke2020responsive}, which is effective at learning constrained policies in a high-dimensional environment with complex reward function but suffers from a slow and unstable training process, and 2) dynamical system modulation (DSM) \cite{huber2019avoidance}, which modulates the nominal policy by multiplying a rotation matrix to avoid infeasible regions. Unlike RL, the modulation requires no explicit training of the policy. Thus, DSM minimizes training time and provides a formal guarantee that the policy will never violate the learned constraints. However, DSM is applicable only for tasks where the control command is the velocity of each state , e.g., controlling the end-effector pose of a robot arm.

\textbf{Constrained reinforcement learning:} A networked policy is trained with a modern constrained RL algorithm PID-Lagrangian\cite{stooke2020responsive}, which offers the advantage of automatically and more stably adjusting the penalty weight. As shown in \eqref{penalized reward}, it reshapes the original reward by incorporating the constraint as a penalty term into the original reward function to avoid the infeasible states. Here, $w_p$ is a penalty weight adjusted by PID-Lagrangian updating rules \cite{stooke2020responsive, peng2022model, peng2021separated}, and $c(s,a)$ is the constraint indicator function. Using this reshaped $r'(s,a)$, a constrained optimal policy can be straightforwardly learned with the standard RL algorithm PPO \cite{schulman2017proximal}.
\begin{equation}
\begin{aligned}
&r'(s,a) = r(s,a) - w_p  c_\theta(s,a), \quad \text{where} \\
 &c_\theta(s,a)=
    \begin{cases}
    0& \text{if $\zeta_\theta(\bar{s}) > d $  (feasible)}  \\
    1& \text{if $\zeta_\theta(\bar{s}) \le d $ (infeasible)} \\
    \end{cases}
\label{penalized reward}
\end{aligned}
\end{equation}

\textbf{Dynamical system modulation:} DSM first learns a nominal unconstrained policy and later modulates nominal policy with the constraint network. The nominal dynamical system policy is represented as \eqref{nominal DS}, where $s_g$ is the goal state and matrix $A(s)$ is a Gaussian Mixture Regression model learned from the known reward function and preserves stability, following \cite{rey2018learning}  \footnote{In our experiment, for DSM baseline we directly set $A(s)=I$ since the task is reaching the destination in the shortest path, which is a straight line connecting the current state and the goal state.}.
\begin{equation}
\begin{aligned}
  &\pi_n(s) \equiv \dot s= A(s)(s-s_g)
\label{nominal DS}
\end{aligned}
\end{equation}

Given a learned constraint network $\zeta_\theta(s)$ with state-only constraint, \eqref{modulated DS} produces a provably safe policy \footnote{The DS modulation theory guarantees that the modulated policy will never violate the learned constraint. Additionally, if the true constraint is a star-shaped obstacle and the nominal policy is linear, the modulated policy also remains globally stable \cite{huber2019avoidance}.  } $\pi(s)$ by multiplying a rotation matrix $M(s)$ constructed from $\zeta_\theta(s)$ following \cite{huber2019avoidance}.
\begin{equation}
\begin{aligned}
  &\pi(s) = M(s)\pi_n(s),  \quad \text{where}\\
  &M(s)=E(s)D(s)E^{-1}(s), \\
  &E(s)=[r(s), e_1(s), e_2(s),... e_{n-1}(s)],  \\
  &n(s) = \frac{d\zeta(s)}{d s}, n(s)\perp e_1(s) \perp e_2(s) \perp ... e_{n-1}(s)\\ 
    &D(s)=diag[\lambda_1(s), \lambda_2(s), \lambda_3(s) ..., \lambda_n(s)], \\
    &\lambda_1(s)=1-\frac{1}{\Gamma(s)}, \lambda_{2,3,...,n}(s)=1+\frac{1}{\Gamma(s)}, \\
    &\Gamma(s)= 1 + 10*(\zeta(s)-0.5)^{0.2} 
\label{modulated DS}
\end{aligned}
\end{equation}

where $n(s)$ is the normal vector to the constraint boundary, $e_i(s)$ are the corresponding tangent vectors, $r(s)$ is the vector towards a reference point inside the obstacle, $\lambda_i(s)$ modifies the velocity along different directions, and $\Gamma(s)$ specifies the relative distance of current state $s$ to the constraint boundary. Intuitively, for any state $s$ close to the constraint boundary, $\Gamma(s)$ approaches 1 and $\lambda_1(s)$  approaches 0, resulting in zero normal velocity and preventing the agent from penetrating the obstacle.

The pseudocode of the proposed method is attached below.

\begin{algorithm}[!htb]
\caption{Positive-Unlabeled Constraint Learning (PUCL)}
\label{alg:ICRL}
\begin{algorithmic}[1] 
\STATE \textbf{input:} Demonstration set $\mathcal{D}_\tau$, randomly initialized constraint network $\zeta_\theta$, randomly initialized policy network $\pi_\phi$, empty memory buffer $\mathcal{M}=\varnothing$
\REPEAT
\STATE \textbf{\% Learning policy}
\STATE Evaluate current policy $\pi_\phi$ and adapt the penalty weight $w_p$ in \eqref{penalized reward} through PID-Lagrangian \cite{stooke2020responsive}
\STATE Update the policy $\pi_\phi$ with the current constraint network $\zeta_\theta$ using constrained RL or DSM
\STATE \textbf{\% Learning constraint with two-step PU learning}
\STATE Sample trajectory set $\mathcal{P}_\tau$ using the current policy $\pi_\phi$ from the same starting points as each demonstration
\STATE Filter sampled set $\mathcal{P}_\tau$ with \eqref{filter condition}: \\
\STATE Identify and expand reliable infeasible set $\mathcal{R}$ using \eqref{RI identification} and \eqref{RI explansion} \\

\STATE Update the constraint $\zeta_\theta$ with the demonstration $\mathcal{D}$, reliable infeasible set $\mathcal{R}$ and memory buffer $\mathcal{M}$ with \eqref{BCE loss}\\

\STATE Save all data from $\mathcal{R}$ into the memory buffer $\mathcal{M}$ 
\UNTIL reach maximum iterations
\end{algorithmic}
\end{algorithm}

\section{Experiment}
\label{sec:experiment}

\subsection{Comparison of Constraint Learning Methods }

To examine the performance of the proposed method, we apply it to four simulated environments to learn a 2D position constraint, a 3D position constraint, a 3D control input constraint and a constraint of an 18D cheetah robot, respectively. All codes are available at \href{https://github.com/epfl-lasa/PUCL_learning_constraint}{https://github.com/epfl-lasa/PUCL\_learning\_constraint.}

\textbf{Baselines:} Four constraint learning methods are compared here, all using constrained RL as the policy: 1) PUCL (the proposed method); 2) maximum-entropy constraint learning (MECL) \cite{anwar2020InverseCR}, a popular IRL-based method that can learn a continuous arbitrary constraint function; 3) binary classifier (BC) \cite{anwar2020InverseCR}, a method directly trains an NN feasibility classifier using loss \eqref{BCE loss} but with all data from $\mathcal{P}$ as infeasible data; 4) generative-PUCL (GPUCL), an alternative PUCL approach using a generative model to identify reliable infeasible data. It first learns a Gaussian Mixture Model (GMM) of the expert distribution, and then identifies reliable infeasible data as the data from $\mathcal{P}$ with the lowest likelihood of being generated by the GMM.  GPUCL is also first formulated in this paper but we do not discuss in detail due to its weaker performance.

\textbf{Metrics:} 
we present the evaluation utilizing two metrics: 1) the IoU (the intersection over union) index, which measures the correctness of the learned constraints. We uniformly sample points in the state space, and compute IoU as (Number of points which are both predicted to be infeasible and truly infeasible) divided by (Number of points either predicted to be infeasible or truly infeasible); the constraint models of all four methods use 0.5 as the decision threshold. 2) the unsafe rate, the per-step true constraint violation rate of the policy learned from the learned constraint.

\textbf{2D/3D Position Constraints:} We first study the two obstacle-avoidance constraints of the end-effector of a robot arm, one in 2D position space and another in 3D position space. We defined a goal-reaching task, where the agent is initialized randomly and rewarded to reach a goal state while staying away from the unknown obstacles. The state space consists of 2D/3D coordinates of the end-effector and the control variables are its corresponding 2D/3D velocities, capped at 0.58m/s. The reward functions in both tasks are the negative of the Cartesian path distance to reach the goal, a natural choice for a goal-reaching task. The constraint networks in both tasks take only the state as input, i.e., $\bar{s}=s$.  

\begin{figure}[!t]
\centering
\subfloat[PUCL (ours)]{\includegraphics[width=0.23\textwidth]{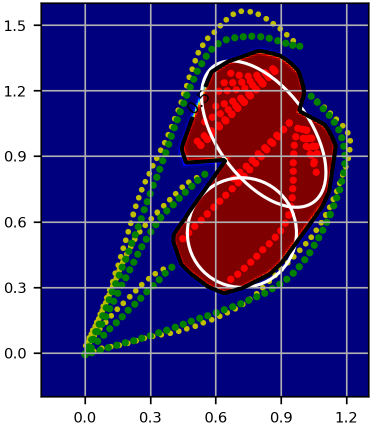}}
\label{fig:2D_pu}
\subfloat[MECL ]{\includegraphics[width=0.23\textwidth]{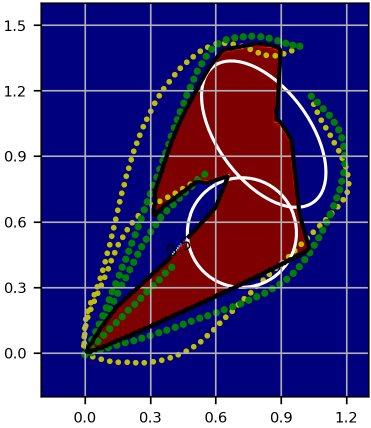}}
\label{fig:2D_ml}

\caption{True constraint and constraint learned with the proposed method PUCL and baseline MECL \cite{anwar2020InverseCR} in the 2D reaching task.  The two white ellipses represent the boundary of the true constraint.  The heat map is a visualization of the learned constraint $\zeta_\theta$, where the red region of $\zeta=0$ represents infeasibility, while the blue regions of $\zeta=1$ indicate feasibility.  The trajectories of the demonstrations and the policy are shown in green and yellow, respectively. The red points correspond to identified reliable infeasible data $\mathcal{R}\cup\mathcal{M}$ (only for PUCL method).}
\label{fig:constraint_visualization}
\end{figure}

In the 2D reaching task, the true constraint involves an irregular obstacle composed of two ellipses as shown in Fig. \ref{fig:constraint_visualization}. In the 3D reaching task, the true constraint is avoiding the tall cylinders. We stress that learning such constraints in our setting is nontrivial and has not been well addressed by prior works. Specifically, our setting does not assume prior knowledge of the obstacles size, shape, or location, while relevant studies \cite{chou2020learningpara} typically require a known parameterization of obstacles and only infer a handful of parameters (e.g., height and radius of cylinders). Furthermore, the studied constraints are nonlinear, not just boxes or hyperplanes, which contrasts with previous works that only learn a plain constraint of a single dimension such as coordinate $x \ge -3$ \cite{anwar2020InverseCR, liu2022benchmarking}. Finally, unlike \cite{chou2022gaussian}, our method does not require a differentiable environmental model.  

The expert demonstration sets for 2D and 3D tasks consist of 4 and 10 safe trajectories, respectively. All demonstrations are generated by an entropy-regularized suboptimal RL agent, trained assuming full knowledge of the true constraint.


A list of important hyper-parameters of our algorithm is given in Table \ref{tab:hyperparam}. For the neural network architectures, the number of hidden units in each layer is mentioned. The hidden activation functions in all networks are Leaky ReLU. The output activation function of the constraint network is the sigmoid function, whereas that of the policy and value network is tanh. Other baselines all adopt identical network architecture.  

\begin{table}[h!]
    \centering
    \caption{List of hyper-parameters of PUCL. }
    \label{tab:hyperparam}
    \begin{tabular}{l|c|c}
    \hline
    & 2D task & 3D task  \\
    \hline
    Policy network & 64, 64 & 64, 64   \\
    Value network & 64, 64 & 64, 64  \\
    Constraint network & 32, 32 & 32, 32  \\
    Policy learning rate & $3 \times 10^{-4}$ & $3 \times 10^{-4}$ \\
    Constraint learning rate & $5 \times 10^{-3}$ & $5 \times 10^{-3}$ \\
    Distance metric & Euclidean & Euclidean   \\
    $\mathcal{R}$ distance threshold $d_r$ & 0.12 & 0.03  \\
    $\mathcal{R}$ identification $k$ & 1 & 1   \\
    sub-optimal coefficient $\delta$ & 0 & 0.03\\
    \hline
    \end{tabular}
\end{table}

\begin{figure}[!t]
\centering
\subfloat{\includegraphics[width=0.19\textwidth]{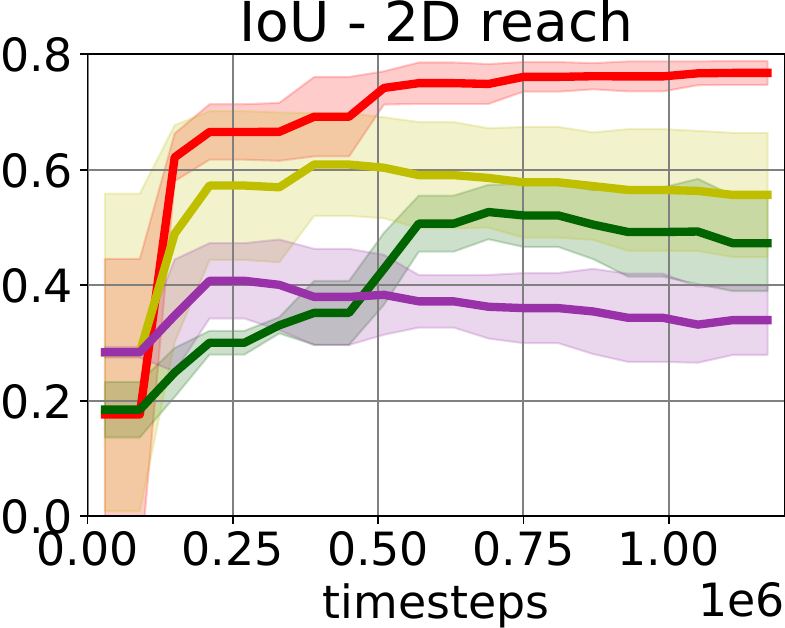}%
\label{fig:2D_jaccard}}
\subfloat{\includegraphics[width=0.19\textwidth]{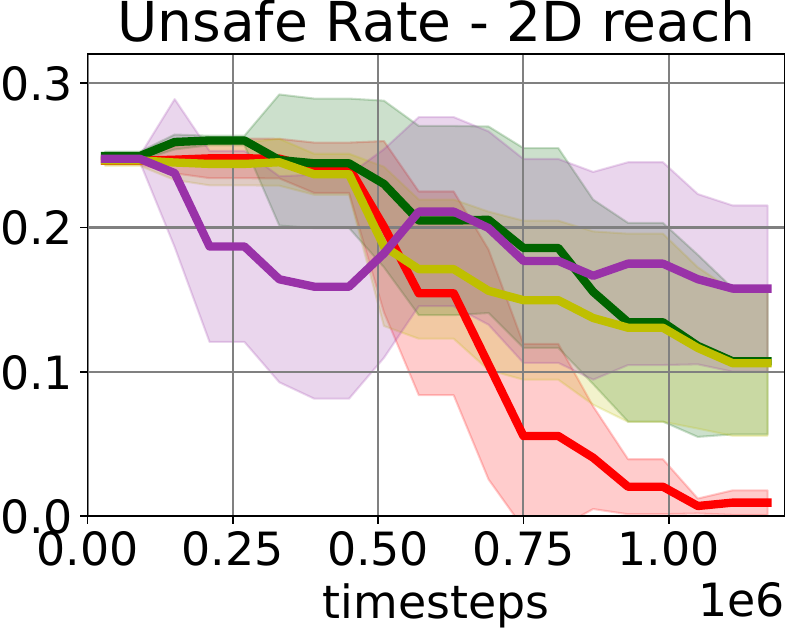}%
\label{fig:2D_violation}}   \\
\subfloat{\includegraphics[width=0.19\textwidth]{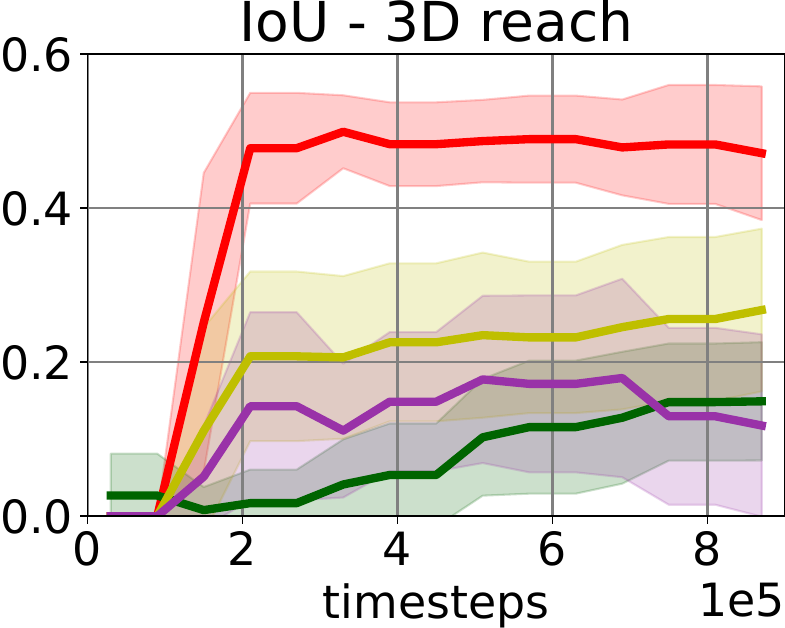}}
\subfloat{\includegraphics[width=0.19\textwidth]{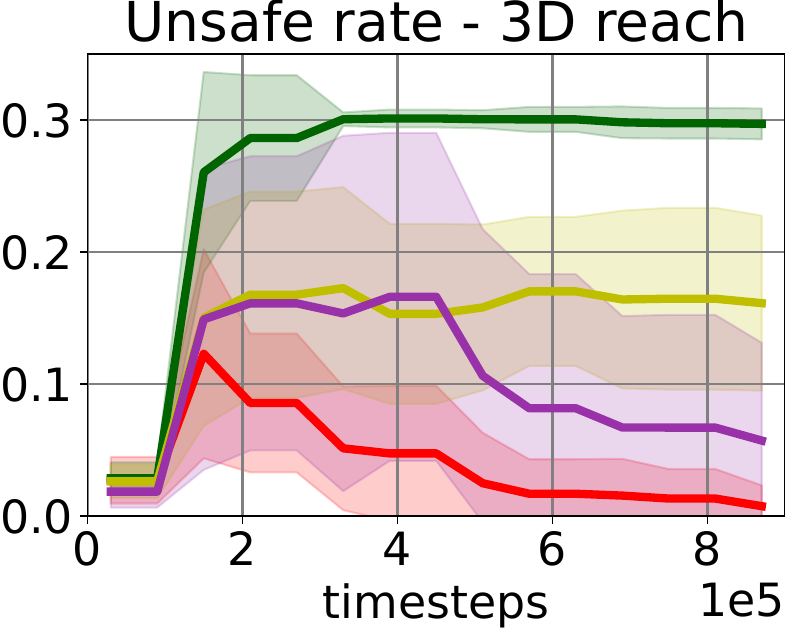}} \\
\subfloat{\includegraphics[width=0.19\textwidth]{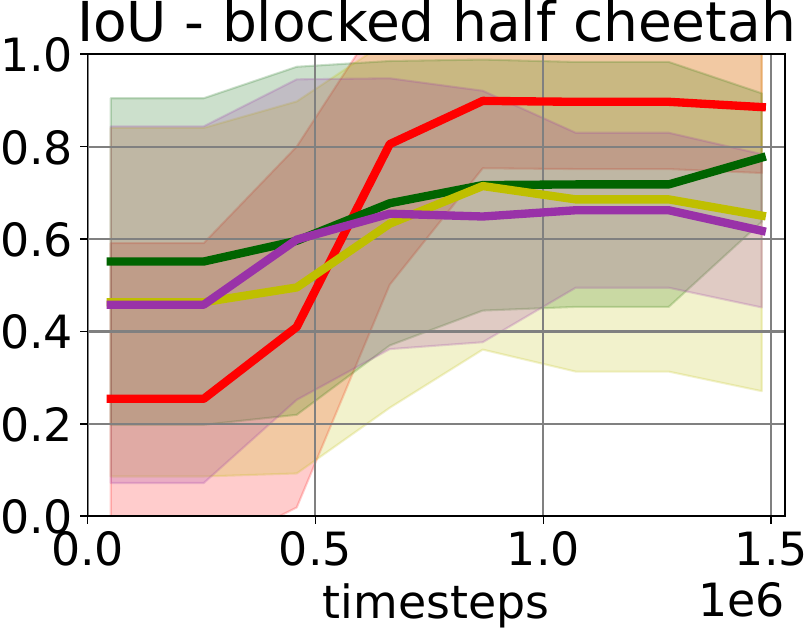}}
\subfloat{\includegraphics[width=0.19\textwidth]{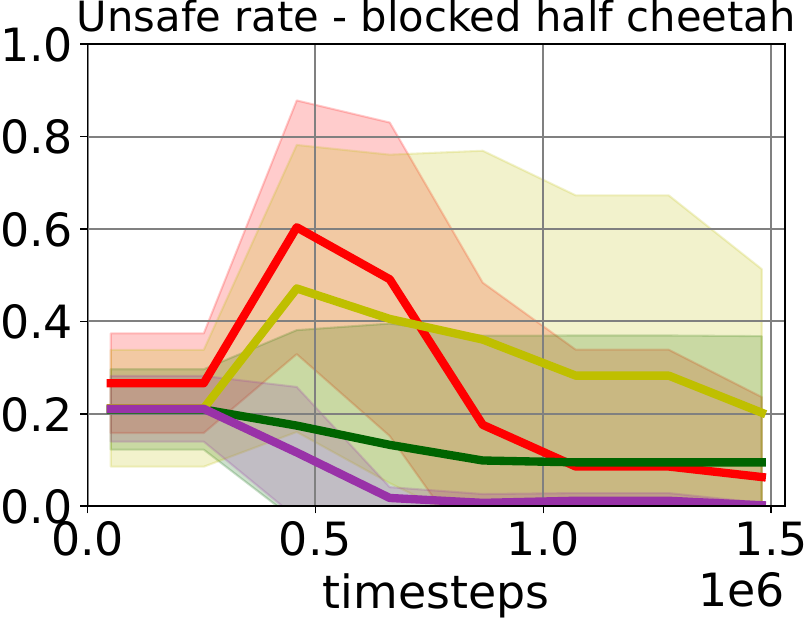}} \\
\subfloat{\includegraphics[width=0.37\textwidth]{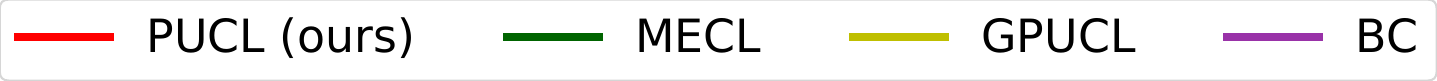}}

\caption{The IoU index (higher is better) and unsafe rate (lower is better) in three environments (top: 2D reaching, middle: 3D reaching, bottom: blocked half cheetah). The x-axis indicates the training process, the number of timesteps the agent takes in the environment.  All the results are the average of 10 independent runs, and the shaded area represents one standard deviation. }
\label{fig:classification_accuracy}
\end{figure}

The IoU index and the unsafe rate in the two tasks are presented in Fig. \ref{fig:classification_accuracy}. Each group of 10 runs uses the same demonstrations and hyper-parameters, but differ in the initialization parameters for the networks. The proposed method PUCL exhibits superior performance compared to the baseline across both environments and metrics. It achieves a near-zero unsafe rate at the convergence of the training in both environments and outperforms baseline MECL. GPUCL adopts a similar framework to PUCL but performs worse than PUCL, possibly due to the difficulty of fitting a good expert distribution with a very limited numbers of demonstrations.  


To provide readers with an intuitive understanding, we include visualizations of the 2D constraint learned with PUCL and MECL in Fig. \ref{fig:constraint_visualization}. Comparing the learned constraints (red region)  with the true constraints (white ellipses) confirms the effectiveness of our method in acquiring a model of a nonlinear constraint. The baseline method MECL learns a less accurate constraint, and even some states visited by the demonstrations are classified as infeasible. 

\begin{figure}[htbp]
\centerline{\includegraphics[width=0.38\textwidth]{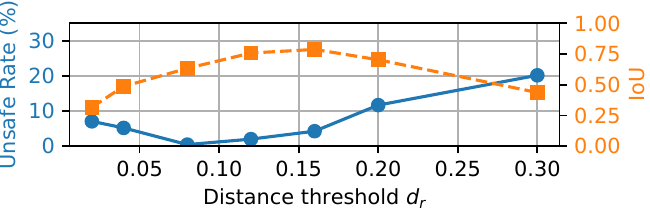}}
\caption{
Sensitivity analysis of the distance threshold $d_r$ on performance. The left y-axis represents the unsafe rate (lower is better), and the right y-axis represents the IoU (higher is better). The curve is averaged on 5 runs with different random seeds.
}\label{fig:sensitivity}
\end{figure}

While our method is effective, it can have a relatively strong dependency on the distance threshold $d_r$. To clearly illustrate the impact of this hyperparameter on performance, we present a sensitivity analysis by varying $d_r$ in the 3D task (see Fig. \ref{fig:sensitivity}).

\textbf{Control Input Constraint:} To demonstrate our method's ability to learn control input constraint, we modify the 3D reaching task by imposing stricter maximum velocity constraints. The original permitted velocities limits are  $|v_x|,|v_y|,|v_z|\leq0.58m/s$.  Imagine that when the human users demonstrate the motion, they may expect the robot to move slower according to their personal preference, e.g., a stricter velocity constraint $|v_x| \leq 0.48m/s, |v_y|\leq0.58m/s, |v_z|\leq 0.19 m/s$. The agent is tasked with inferring this user-specific velocity constraint from demonstrations. Importantly, although the true constraints follow a box form, the agent is not given any prior knowledge of this structure; it cannot assume a box constraint and simply find maximum values from the data. For simplicity, the obstacle-avoidance constraint is assumed to be known in this task, and only the velocity constraint is learned, thus $\bar{s}=a$. 



\begin{figure}[htbp]
\centerline{\includegraphics[width=0.45\textwidth]{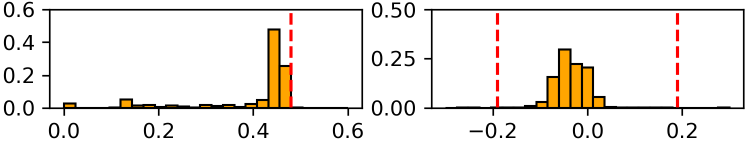}}
\caption{Velocity distribution of the  policy  learned with PUCL (Left: $v_x$, Right: $v_z$). The horizontal axis represents the velocity, and the red line represents the true constraint threshold.}
\label{fig:rechvel hist}
\end{figure}

We collected 30 demonstrations from an expert RL agent and applied PUCL and MECL to learn the velocity constraint. For PUCL, we used the Euclidean distance with a threshold of $d_r = 0.01$. The results, averaged over 8 runs, showed that PUCL achieved an IoU score of $0.87 \pm 0.08$, compared to $0.77 \pm 0.26$ for MECL. In terms of unsafe rate, PUCL achieves an unsafe rate of only $(3.47 \pm 3.26)\%$ compared to $(28.1 \pm 19.2)\%$ for MECL. Furthermore, the velocity distribution output by the policy learned with PUCL is shown in Fig.~\ref{fig:rechvel hist}, demonstrating that almost all data points fall within the true safe range.

\textbf{Constraint in high-dimensional space:} The previous tasks extracts constraints only from 2D/3D spaces. To showcase our method extends to high-dimensional, heterogeneous state space, we apply it to an 18D blocked half cheetah environment introduced by baseline papers \cite{anwar2020InverseCR, liu2022benchmarking}. The agent controls a cheetah robot with two legs by applying torques on each joint motors. The reward is determined by the distance it walks between the current and the previous time step. The constraint blocks the region with x-coordinate $\le-3$, so the robot is only allowed to move in the region with x-coordinate between $-3$ and $\infty$. The input to the constraint network is 18-dimensional state space, including positional and angular values of different body parts of the cheetah, followed by their velocities, making constraint extraction challenging due to this high-dimensional space.

We collected 30 demonstrations and run PUCL to learn constraint. Since each dimension of the state space represents distinct physical properties and scales, we identify the reliable infeasible data based on the standardized dataset as discussed in \ref{sec:method pucl}. Euclidean distance is used as the distance measure \footnote{We also tested Manhattan distance and Chebyshev distance. Both achieves similar performance with appropriately selected distance thresholds.}  with a threshold  \footnote{This value is tuned following the discussion in \ref{sec:method pucl} for heterogeneous cases: we first collect $\mathcal{P}$ using an unconstrained policy and compute the score $\mathcal{S}_\mathcal{D}(\bar{s})$ of each state from $\mathcal{P}$. We hypothesis that around 80\% of states in $\mathcal{P}$ are reliable infeasible data, and this corresponds to a threshold of 2.6.}  $d_r=2.6$. The learning performance is presented in Fig. \ref{fig:classification_accuracy}, where PUCL achieves the highest IoU metrics
and the second best unsafe rate. This confirms the ability of PUCL for learning constraints in high-dimensional state spaces.

\subsection{Comparison of Policy Representation and Learning Methods}
\label{sec: different policy learning results}
As introduced in section \ref{sec:policy learning}, we studied two policy representation and learning approaches, namely constrained RL and DSM. We test the two variants with PUCL as the constraint learning method in the 3D reaching task. The two frameworks are named CRL-PUCL and DSM-PUCL, respectively, and the results of comparison are presented in Table \ref{tab:different policy learning}. \footnote{ The recall rate is computed as (True infeasible)/(True infeasible + False feasible), and precision is (True infeasible)/(True infeasible + False infeasible) } Training was conducted on a PC with a 12th Intel i9-12900K × 24 and an NVIDIA GeForce RTX 3070. All results are the average of 15 independent runs. 

DSM-PUCL and CRL-PUCL achieve similar classification performance in IoU, recall rate and precision, and DSM-PUCL generally exhibits lower variance. For the learned policy, while both methods have an unsafe rate below $1\%$, the rate of DSM-PUCL is slightly lower than CRL-PUCL. This is because DSM-PUCL modulates the policy in a principled way and features a safety guarantee \cite{huber2019avoidance}. DSM-PUCL also requires less training time than CRL-PUCL. Based on these comparisons, we conclude that both methods are effective for learning constraints. For robotic tasks where the control variable is the velocity/acceleration of each state dimension, DSM may be more favourable than CRL in terms of training time and policy performance, while CRL is more suitable for more general task with more complex environment dynamics and rewards.

\begin{table}[H]
\caption{Comparison of two policy representation and learning methods}
\begin{center}
\begin{tabular}{lccccc}
\hline
Methods     & CRL-PUCL   & DSM-PUCL                      \\ \hline
IoU   & $0.46 \pm 0.09$     & $0.42 \pm 0.01$           \\ \hline
Recall rate   & $(94.2 \pm 4.5)$\%     & $(99.9 \pm 0.1)$ \%        \\ \hline
Precision   & $(47.6 \pm 10.0)$\%      & $(42.1 \pm 1.3$) \%           \\ \hline
Unsafe rate  & $(0.47 \pm 0.82)$\%      & $(0.15 \pm 0.56)$\%             \\ \hline
Training time (min)   &$ 13.4 \pm 1.7$     & $4.7 \pm 0.4$          \\ \hline
\end{tabular}
\end{center}
\label{tab:different policy learning}
\end{table}

\subsection{Constraint Transfer to Variant of the Same Task}
\label{sec: constraint transfer}
One remarkable advantage of learning constraint is that the learned constraint can be transferred to relevant task with the same constraint but potentially different goals and rewards. To demonstrate this, we consider the task of robot avoiding four cups forming two sets of complex shapes (see Fig. \ref{fig:panda cups}). We collected demonstrations using DSM and applied DSM-PUCL to learn this constraint. 
Then the learned constraint network is transferred to a variant task with a set of shifted goal states and starting points. The new nominal policy heading to new target is directly modulated with the transferred constraint with DSM. The modulated policy also avoids these cups despite the change of task configuration. 

\section{Conclusions and Limitations}
\label{sec:conclusion}
This paper proposed a two-step Positive-Unlabeled Constraint Learning method to infer an arbitrary, potentially nonlinear, constraint function from demonstration. The proposed method treats the demonstration as positive data and the higher-reward policy as unlabeled data. It uncovers the infeasible states by first identifying reliable infeasible data using a distance metric, and training a feasibility classifier with the identified data. The benefits of the proposed method were demonstrated by learning a 2D and 3D position constraint, a 3D velocity constraint and a constraint in high-dimensional environment, using either a constrained RL policy or a DSM policy. It managed to recover the constraints and outperformed other baseline methods in terms of accuracy and safety. 

Although the proposed method is effective, its performance depends on the choice of an appropriate distance threshold and the specific distance metric. In the future, we aim to address this limitation by exploring methods to jointly learn a distance representation that can automatically adapt to various tasks.






\bibliographystyle{IEEEtran}
\bibliography{ref}

\end{document}